\documentclass[final]{cvpr}

\usepackage{times}
\usepackage{epsfig}
\usepackage{graphicx}
\usepackage{comment}
\usepackage{amsmath,amssymb} 
\usepackage{color}
\usepackage{subcaption}
\usepackage[dvipsnames]{xcolor}
\usepackage{booktabs}
\usepackage{multirow}
\usepackage[flushleft]{threeparttable}

\usepackage{stmaryrd}

\newcommand{\omitme}[1]{}
\usepackage[pagebackref=true,breaklinks=true,colorlinks,bookmarks=false]{hyperref}

\def\cvprPaperID{6542} 
\def\confYear{CVPR 2021}

\begin{document}

\title{MobileDets: Searching for Object Detection Architectures for Mobile Accelerators}

\author{Yunyang Xiong\thanks{Equal contribution.}\\
University of Wisconsin-Madison\\
{\tt\small yxiong43@wisc.edu}
\and
Hanxiao Liu$^{\ast}$\\
Google\\
{\tt\small hanxiaol@google.com}
\and
Suyog Gupta\\
Google\\
{\tt\small suyoggupta@google.com}
\and
Berkin Akin, Gabriel Bender, Yongzhe Wang, Pieter-Jan Kindermans, Mingxing Tan \\
Google \\
{\tt\small \{bakin,gbender,yongzhe,pikinder,tanmingxing\}@google.com}
\and
Vikas Singh\\
University of Wisconsin-Madison\\
{\tt\small vsingh@biostat.wisc.edu}
\and
Bo Chen \\
Google \\
{\tt\small bochen@google.com}
}
\maketitle

\begin{abstract}
\vspace{-3pt}
Inverted bottleneck layers, which are built upon depthwise convolutions, have been the predominant building blocks in state-of-the-art object detection models on mobile devices. In this work, we investigate the optimality of this design pattern over a broad range of mobile accelerators by revisiting the usefulness of regular convolutions.
We discover that regular convolutions are a potent component to boost the latency-accuracy trade-off for object detection on accelerators, provided that they are placed strategically in the network via neural architecture search. By incorporating regular convolutions in the search space and directly optimizing the network architectures for object detection, we obtain a family of object detection models, MobileDets, that achieve state-of-the-art results across mobile accelerators.
On the COCO object detection task, MobileDets outperform MobileNetV3+SSDLite by $1.7$ mAP at comparable mobile CPU inference latencies.
MobileDets also outperform MobileNetV2+SSDLite by $1.9$ mAP on mobile CPUs, $3.7$ mAP on Google EdgeTPU, $3.4$ mAP on Qualcomm Hexagon DSP and $2.7$ mAP on Nvidia Jetson GPU without increasing latency.
Moreover, MobileDets are comparable with the state-of-the-art MnasFPN on mobile CPUs even without using the feature pyramid, and achieve better mAP scores on both EdgeTPUs and DSPs with up to $2\times$  speedup. Code and models are available in the TensorFlow Object Detection API~\cite{huang2017speed}: \url{https://github.com/tensorflow/models/tree/master/research/object_detection}.
\end{abstract}

\section{Introduction}
In many computer vision applications it can be observed that higher capacity networks lead to superior performance \cite{szegedy2017inception,zoph2018learning,hu2018squeeze,real2019regularized}. However, they are often more resource-consuming.
This makes it challenging to find models with the right quality-compute trade-off for deployment on edge devices with limited inference budgets.

A lot of effort has been devoted to the manual design of lightweight neural architectures for edge devices~\cite{howard2017mobilenets,sandler2018mobilenetv2,zhang2018shufflenet,xiong2019antnets}. Unfortunately, relying on human expertise is time-consuming and can be sub-optimal. This problem is made worse by the speed at which new hardware platforms are released. In many cases, these newer platforms have differing performance characteristics which make a previously developed model sub-optimal.

To address the need for automated tuning of neural network architectures, many methods have been proposed. In particular, neural architecture search (NAS) methods \cite{cai2018proxylessnas,tan2019efficientnet,tan2019mnasnet,howard2019searching,dai2019chamnet} have demonstrated a superior ability in finding models that are not only accurate but also efficient on a specific hardware platform.

Despite many advancements in NAS algorithms~\cite{zoph2016neural,real2019regularized,bender2018understanding,liu2018darts,cai2018proxylessnas,tan2019efficientnet,xiong2019resource,tan2019mnasnet,howard2019searching,dai2019chamnet}, it is remarkable that inverted bottlenecks (IBN) \cite{sandler2018mobilenetv2} remain the predominant building block in state-of-the-art mobile models. IBN-only search spaces have also been the go-to setup in a majority of the related NAS publications \cite{tan2019mnasnet,cai2018proxylessnas,tan2019efficientnet,howard2019searching}. 
IBN layers rely heavily on depthwise and depthwise separable convolutions \cite{depconv14}. The resulting models have relatively low FLOPS and parameter counts, and can be executed efficiently on CPUs. 

However, the advantage of depthwise convolutions for mobile inference is less clear for hardware accelerators such as DSPs or EdgeTPUs
which are becoming increasingly popular on mobile devices.
For example, for certain tensor shapes and kernel dimensions, a regular convolution can run $3\times$ as fast as the depthwise variation on an EdgeTPU despite having $7\times$ as many FLOPS. This observation leads us to question the exclusive use of IBN-only search spaces in most current state-of-the-art mobile architectures.

Our work seeks to rethink the use of IBN-only search space on modern mobile accelerators. To this end, we propose the MobileDet search space family, which includes not only IBNs but also flexible full convolution sequences motivated by the structure of tensor decomposition \cite{tucker1966some,carroll1970analysis,mehta2019scaling}.
Using the task of object detection as an example (one of the most popular mobile vision applications), we show the MobileDet search space family enables NAS methods to identify models with substantial better latency-accuracy trade-offs on mobile CPUs, DSPs, EdgeTPUs and edge GPUs.

\begin{figure*}[h]
\centering
    \includegraphics[width=0.65\linewidth]{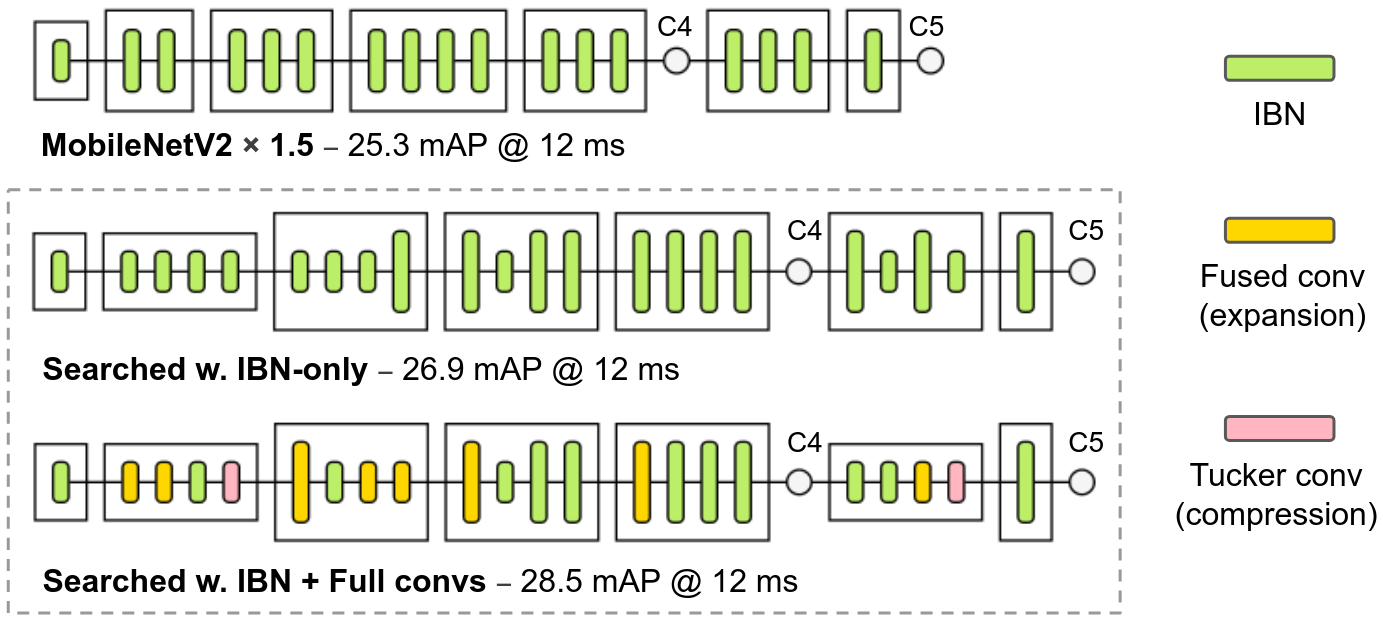}
    \caption{\footnotesize Platform-aware NAS and our MobileDet search space work synergistically to boost object detection performance on accelerators. SSDLite object detection performance on Pixel-4 DSPs with different backbone designs: manually-designed MobileNetV2, searched with IBN-only search space, and searched with the proposed MobileDet space (with both IBNs and full conv-based building blocks). Layers are visualized as vertical bars where color indicates layer type and length indicates expansion ratio. $C4$ and $C5$ mark the feature inputs to the SSDLite head. While conducting platform-aware NAS in an IBN-only search space achieves a $1.6$ mAP boost over the handcrafted baseline, searching within the MobileDet space brings another $1.6$ mAP gain.}
    \label{fig:architecture-comparisons}
\end{figure*}

To evaluate our proposed MobileDet search space, we perform latency-aware NAS for object detection,
targeting a diverse set of mobile platforms.
Experimental results show that by using our MobileDet search space family and directly searching on detection tasks, we can consistently improve the performance across all hardware platforms. By leveraging full convolutions at selected positions in the network, our method outperforms IBN-only models by a significant margin. Our searched models, MobileDets,
outperform MobileNetV2 classification backbone by 1.9mAP on CPU, 3.7mAP on EdgeTPU, 3.4mAP on DSP, and 2.7mAP on edge GPU at comparable inference latencies.
MobileDets also outperform the state-of-the-art MobileNetV3 classification backbone by 1.7mAP at similar CPU inference efficiency.
Further,
the searched models achieved comparable performance with the state-of-the-art mobile CPU detector,
MnasFPN~\cite{chen2019mnasfpn}, without leveraging the NAS-FPN head which may complicate the deployment.
On both EdgeTPUs and DSPs, MobileDets are more accurate than MnasFPN while being more than twice as fast.

Our main contributions can be summarized as follows:
\begin{itemize}
    \item Unlike many existing works which exclusively focused on IBN layers for mobile applications,
    we propose an augmented search space family with building blocks based on regular convolutions.
    We show that NAS methods can substantially benefit from this enlarged search space to achieve better latency-accuracy trade-off on a variety of mobile devices.
    \item We deliver MobileDets, a set of mobile object detection models with state-of-the-art quality-latency trade-offs on multiple hardware platforms, including mobile CPUs, EdgeTPUs, DSPs and edge GPUs. Code and models will be released to benefit a wide range of on-device object detection applications.
\end{itemize}

\section{Related Work}
\subsection{Mobile Object Detection}
 Object detection is a classic computer vision challenge where the goal is to learn to identify objects of interest in images. Existing object detectors can be divided into two categories: two-stage detectors and one-stage single shot detectors. For two-stage detectors, including Faster RCNN \cite{ren2015faster}, R-FCN \cite{dai2016r} and ThunderNet \cite{qin2019thundernet}, region proposals must be generated first before the detector can make any subsequent predictions. Two-stage detectors are not efficient in terms of inference time due to this multi-stage nature. On the other hand, one-stage single shot detectors, such as SSD \cite{liu2016ssd}, SSDLite \cite{sandler2018mobilenetv2}, YOLO \cite{redmon2016you}, SqueezeDet \cite{wu2017squeezedet} and Pelee \cite{wang2018pelee}, require only one single pass through the network to predict all the bounding boxes, making them ideal candidates for efficient inference on edge devices. We therefore focus on one-stage detectors in this work.

SSDLite \cite{sandler2018mobilenetv2} is an efficient variant of SSD that has become one of the most popular lightweight detection heads.
It is well suited for use cases on mobile devices.
Efficient backbones, such as MobileNetV2 \cite{sandler2018mobilenetv2} and MobileNetV3 \cite{howard2019searching}, are paired with SSDLite to achieve state-of-the-art mobile detection results.
Both models will be used as baselines to demonstrate the effectiveness of our proposed search spaces over different mobile accelerators.

\subsection{Mobile Neural Architecture Search (NAS)}

 NetAdapt~\cite{yang2018netadapt} and AMC~\cite{he2018amc} were among the first attempts to utilize latency-aware search to finetune the number of channels of a pre-trained model.
MnasNet~\cite{tan2019mnas} and MobileNetV3~\cite{howard2019searching} extended this idea to find resource-efficient architectures within the NAS framework.
With a combination of techniques, MobileNetV3 delivered state-of-the-art architectures on mobile CPU. As a complementary direction,
there are many recent efforts aiming to improve the search efficiency of NAS~\cite{brock2017smash,bender2018understanding,pham2018efficient,liu2018darts,cai2018proxylessnas,wu2019fbnet,cai2019once}.

\subsection{NAS for Mobile Object Detection}
A large majority of the NAS literature \cite{tan2019efficientnet,tan2019mnasnet,howard2019searching} focuses on classification and only re-purposes the learned feature extractor as the backbone for object detection without further searches. Recently, multiple papers \cite{chen2019detnas,wang2019fcos,chen2019mnasfpn} have shown that better latency-accuracy trade-offs are obtained by searching directly for object detection models. 

One strong detection-specific NAS baseline for mobile detection models is MnasFPN~\cite{chen2019mnasfpn}, which searches for the feature pyramid head with a mobile-friendly search space that heavily utilizes depthwise separable convolutions. Several factors limit its generalization towards mobile accelerators: (1) so far both depthwise convolutions and feature pyramids are less optimized on these platforms, and (2) MnasFPN does not search for the backbone, which is a bottleneck for latency. By comparison, our work relies on SSD heads and proposes a new search space for the backbone based on full-convolutions, which are more amenable to mobile acceleration.
While it is challenging to develop a generic search space family that spans a set of diverse and dynamically-evolving mobile platforms, we take a first step towards this goal, starting from the most common platforms such as mobile CPUs, DSPs and EdgeTPUs.

\section{Revisiting Full Convolutions for Mobile Search Spaces}
In this section, we first explain why IBN layers may not be sufficient to handle mobile accelerators beyond mobile CPUs. We then propose new building blocks based on regular convolutions to enrich our search space,
and discuss the connections between these building blocks and Tucker/CP decompositions~\cite{tucker1966some,carroll1970analysis}.

\begin{figure}[h!]\centering
	\includegraphics[width=0.8\columnwidth]{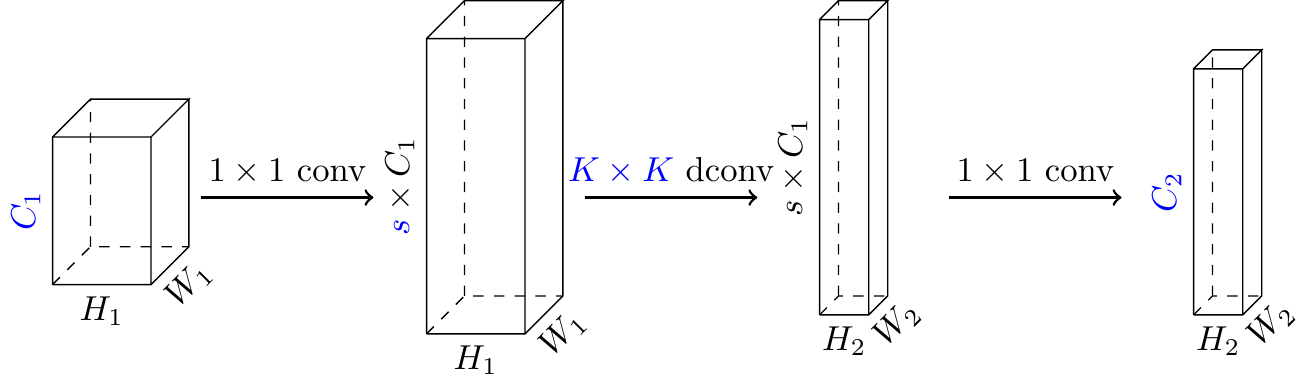}
	\vspace{-.8em}
	\caption{\label{fig:cp_layers}\footnotesize Inverted bottleneck layer: $1\times 1$ pointwise convolution transforms the input channels from $C_1$ to $s\times C_1$ with input expansion ratio $s > 1$, then $K\times K$ \textit{depthwise} convolution transforms the input channels from $s\times C_1$ to $s \times  C_1$, and the last $1\times 1$ pointwise convolution transforms the channels from $s \times C_1$ to $C_2$. The highlighted $C_1, s, K, C_2$ in IBN layer are searchable.}
\end{figure}

\paragraph{Are IBNs all we need?} The layout of an Inverted Bottleneck (IBN) is illustrated in Figure~\ref{fig:cp_layers}.
IBNs are designed to reduce the number of parameters and FLOPS, and leverage depthwise and pointwise (1x1) convolutional kernels to achieve high efficiency on mobile CPUs. However, not all FLOPS are the same, especially for modern mobile accelerators such as EdgeTPU and DSPs.
For example, a regular convolution may run $3\times$ as fast on EdgeTPUs than its depthwise variation even with $7\times$ as many FLOPS. The observation indicates that the widely used IBN-only search space can be suboptimal for modern mobile accelerators.
This motivated us to propose new building blocks by revisiting regular (full) convolutions to enrich IBN-only search spaces for mobile accelerators.
Specifically,
we propose two flexible layers to perform channel expansion and compression, respectively, which are detailed below.

\begin{figure}[h!]\centering
	\includegraphics[width=0.8\columnwidth]{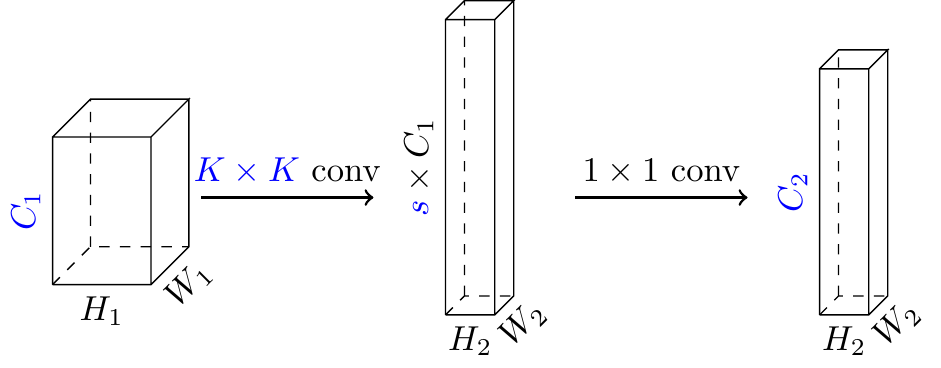}
	\vspace{-.8em}
	\caption{\label{fig:fused_layers}\footnotesize Fused inverted bottleneck layer: $K\times K$ \textit{regular} convolution transforms the input channels from $C_1$ to $s \times C_1$ with input expansion ratio $s > 1$, and the last $1\times 1$ pointwise convolution transforms the channels from $s \times C_1$ to $C_2$. The highlighted $C_1, K, s, C_2$ in the fused inverted bottleneck layer are searchable.}
\end{figure}

\begin{figure}[h!]\centering
	\includegraphics[width=0.95\columnwidth]{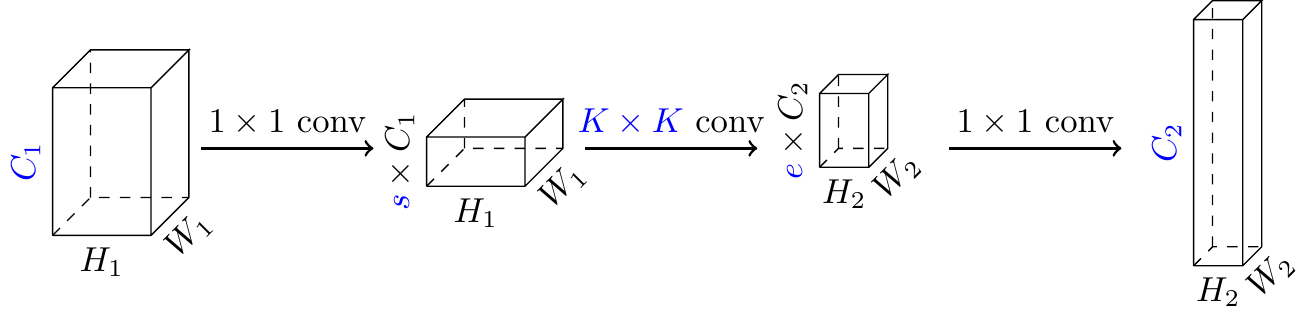}
	\vspace{-.8em}
	\caption{\label{fig:tucker_layers}\footnotesize Tucker layer: $1\times 1$ pointwise convolution transforms the input channels $C_1$ to $s\times C_1$ with input compression ratio $s < 1$, then $K\times K$ \textit{regular} convolution transforms the input channels from $s\times C_1$ to $e \times C_2$ with output compression ratio $e < 1$, and the last $1\times 1$ pointwise convolution transforms the channels from $e \times C_2$ to $C_2$. The highlighted $C_1, s, K, e, C_2$ in Tucker layer are searchable.}
\end{figure}

\subsection{Fused Inverted Bottleneck Layers (Expansion)}
The depthwise-separable convolution~\cite{depconv14} is a critical element of an inverted bottleneck~\cite{howard2017mobilenets} (Figure~\ref{fig:cp_layers}). The idea behind the depthwise-separable convolution  is to replace an ``expensive'' full convolution with a combination of a depthwise convolution (for spatial dimension) and a $1\times 1$ pointwise convolution (for channel dimension).
However,
the notion of expensiveness was largely defined based on FLOPS or the number of parameters,
which are not necessarily correlated with the inference efficiency on modern mobile accelerators. To incorporate regular convolutions, we propose to modify an IBN layer by fusing together its first $1\times1$ convolution and its subsequent $K \times K$ depthwise convolution into a single $K \times K$ regular convolution (Figure~\ref{fig:fused_layers}).
Like a standard inverted bottleneck layer, the initial convolution in our fused inverted bottleneck increases the number of filters by a factor $s > 1$.
The expansion ratio of this layer will be determined by the NAS algorithm.
\subsection{Tucker Convolution Layers (Compression)}
Bottleneck layers were introduced in ResNet~\cite{he2016deep} reduce the cost of large convolutions over high-dimensional feature maps. A bottleneck layer with compression ratio $s < 1$ consists of a $1 \times 1$ convolution with $C_1$ input filters and $s \cdot C_1$ output filters, followed by a $K \times K$ convolution with $s \cdot C_1$ output filters, followed by a $1 \times 1$ convolution with $C_2$ output filters. We generalize these bottlenecks (Figure \ref{fig:tucker_layers}) by allowing the initial $1 \times 1$ convolution to potentially have a different number of output filters than the $K \times K$ convolution and let the NAS algorithm to decide the best configurations.

We refer to these new building blocks as \emph{Tucker convolution layers} because of their connections with Tucker decomposition. (See appendix for details.)
\section{Architecture Search Method}
\paragraph{Search Algorithm.}
Our proposed search spaces are complementary to any neural architecture search algorithms.
In our experiments,
we employ TuNAS~\cite{bender2020can} for its scalability and its reliable improvement over random baselines. TuNAS constructs a one-shot model that encompasses all architectural choices in a given search space, as well as a controller whose goal is to pick an architecture that optimizes a platform-aware reward function. The  one-shot model and the controller are trained together during a search. In each step, the controller samples a random architecture from a multinomial distribution that spans over the choices, then the portion of the  one-shot model's weights associated with the sampled architecture are updated, and finally a reward is computed for the sampled architecture, which is used to update the controller. The update is given by applying REINFORCE algorithm \cite{williams1992reinforce} on the following reward function:

\begin{equation*} \label{eq:unconopt}
R(M) = \textit{mAP}(M)+\tau\left|\frac{c(M)}{c_0}-1\right|
\end{equation*}
where $\textit{mAP}(M)$ denotes the mAP of an architecture $M$, $c(M)$ is the inference cost (in this case, latency), $c_0$ is the given cost budget, and $\tau < 0$ is a hyper-parameter that balances the importance of accuracy and inference cost. The search quality tends to be insensitive to $\tau$, as shown in~\cite{bender2020can}.

\paragraph{Cost Models.}
We train a \emph{cost model}, $c(\cdot)$ -- a linear regression model whose features are composed of, for each layer, an indicator of the cross-product between input/output channel sizes and layer type. This model has high fidelity across platforms ($r^2\geq 0.99$). The linear cost model is related to previously proposed methods based on lookup tables~\cite{yang2018netadapt,he2018amc,tan2019mnasnet}, but only requires us to benchmark the latencies of randomly selected models within the search space, and does not require us to measure the costs of individual network operations such as convolutions.

Since $R(M)$ is computed at every update step, efficiency is key. During search, we estimate $mAP(M)$ based on a small mini-batch for efficiency, and use the regression model as a surrogate for on-device latency $c(M)$. To collect training data for the cost model, we randomly sample several thousand network architectures from our search space and benchmark each one on device. This is done only once per hardware and prior to search, eliminating the need for direct communication between server-class ML hardware and mobile phones. For final evaluation, the found architectures are benchmarked on the actual hardware instead of the cost model.

\section{Experiments}
We use the COCO dataset for our object detection experiments. We report mean average precision (mAP) for object detection and the real latency of searched models on accelerators with image size 320$\times$320. We conduct our experiments in two stages: architecture search for optimal architecture and architecture evaluation via retraining the found architecture from scratch. 

\subsection{Implementation Details}
\noindent\textbf{Standalone Training}. 
We use 320$\times$320 image size for both training and evaluation.
The training is carried out using 32 Cloud TPU v2 cores.
For fair comparison with existing models, we use standard preprocessing in the Tensorflow object detection API without additional enhancements such as drop-block or auto-augment.
We use SGD with momentum 0.9 and weight decay $5\times 10^{-4}$. The learning rate is warmed up in the first 2000 steps and then follows cosine decay.
All models are trained from scratch without any ImageNet pre-trained checkpoint.
We consider two different training schedules:
\begin{itemize}
    \item \emph{Short-schedule}: Each model is trained for 50K steps with batch size 1024 and an initial learning rate of 4.0.
    \item \emph{Long-schedule}: Each model is trained for 400K steps with batch size 512 and an initial learning rate of 0.8.
\end{itemize}
The short schedule is about $4\times$ faster than the long schedule but would result in slightly inferior quality.
Unless otherwise specified, we use the short schedule for ablation studies and the long schedule for the final results in Table~\ref{tab:main-results}.

\smallskip
\noindent\textbf{Architecture Search}. 
To avoid overfitting the true validation dataset, we split out 10\% of the COCO training data to evaluate the models and compute rewards during search.
Hyperparameters for training the shared weights follow the short schedule for standalone training.
As for reinforcement learning, we use Adam optimizer with an initial learning rate of $5\times 10^{-3}$, $\beta = (0, 0.999)$ and $\epsilon = 10^{-8}$.
We search for 50K steps for ablation studies and search for 100K steps to obtain the best candidates in the main results table.

\subsection{Latency Benchmarking}
We report on-device latencies for all of our main results.
We benchmark using TF-Lite for CPU, EdgeTPU and DSP, which relies on NNAPI to delegate computations to accelerators. For all benchmarks, we use single-thread and a batch size of $1$. In Pixel 1 CPU, we use only a single large core. For Pixel 4 EdgeTPU and DSP, the models are fake-quantized~\cite{jacob2018quantization} as required. The GPU models are optimized and benchmarked using TensorRT 7.1 converted from an intermediate ONNX format.

\subsection{Search Space Definitions}
The overall layout of our search space resembles that of ProxylessNAS~\cite{cai2018proxylessnas} and TuNAS~\cite{bender2020can}.
We consider three variants with increasing sizes:
\begin{itemize}
    \item \emph{IBN}: The smallest search space that contains IBN layers only. Each layer may choose from kernel sizes $(3,5)$ and expansion factors $(4,8)$.
    \item \emph{IBN+Fused}: An enlarged search space that not only contains all the IBN variants above,
    but also Fused convolution layers in addition with searchable kernel sizes $(3,5)$ and expansion factors $(4,8)$.
    \item \emph{IBN+Fused+Tucker}: A further enlarged  search space that contains Tucker (compression) layers in addition. Each Tucker layer allows searchable input and output compression ratios within $(0.25, 0.75)$.
\end{itemize}
For all search spaces variants above,
we also search for the output channel sizes
of each layer among the options of $(0.5, 0.625, 0.75, 1.0, 1.25, 1.5, 2.0)$ times a \emph{base channel size} (rounded to multiples of 8 to be more hardware-friendly).
Layers in the same block share the same base channel size, though they can end up with different expansion factors. The base channel sizes for all the blocks (from stem to head) are 32-16-32-48-96-96-160-192-192.
The multipliers and base channel sizes are designed to approximately subsume several representative architectures in the literature, such as MobileNetV2 and MnasNet.

\smallskip
\noindent\textbf{Hardware-Specific Adaptations}. 
The aforementioned search spaces are slightly adapted depending on the target hardware. Specifically: \textbf{(a)} all building blocks are augmented with Squeeze-and-Excitation blocks and h-Swish activation functions (to replace ReLU-6) when targeting CPUs.
    This is necessary to obtain a fair comparison against the MobileNetV3+SSDLite baseline, which also includes these operations. Neither primitive is well supported on EdgeTPUs or DSPs; \textbf{(b)} when targeting DSPs, we exclude 5$\times$5 convolutions from the search space, since they are highly inefficient due to hardware/software constraints.

\subsection{Search Space Ablation}
\label{sec:search-space-ablation}
With a perfect architecture search algorithm,
the largest search space is guaranteed to outperform the smaller ones because it subsumes the solutions of the latter. This is not necessarily the case in practice, however, as the algorithm may end up with sub-optimal solutions, especially when the search space is large~\cite{chen2019mnasfpn}.
In this work,
a search space is considered useful if it enables NAS methods to identify \emph{sufficiently good} architectures even if they are not optimal.
In the following,
we evaluate the usefulness of different search spaces by pairing them with TuNAS \cite{bender2020can}, a scalable latency-aware NAS implementation.

\begin{figure*}[!t]
    \centering
    \includegraphics[width=0.75\linewidth]{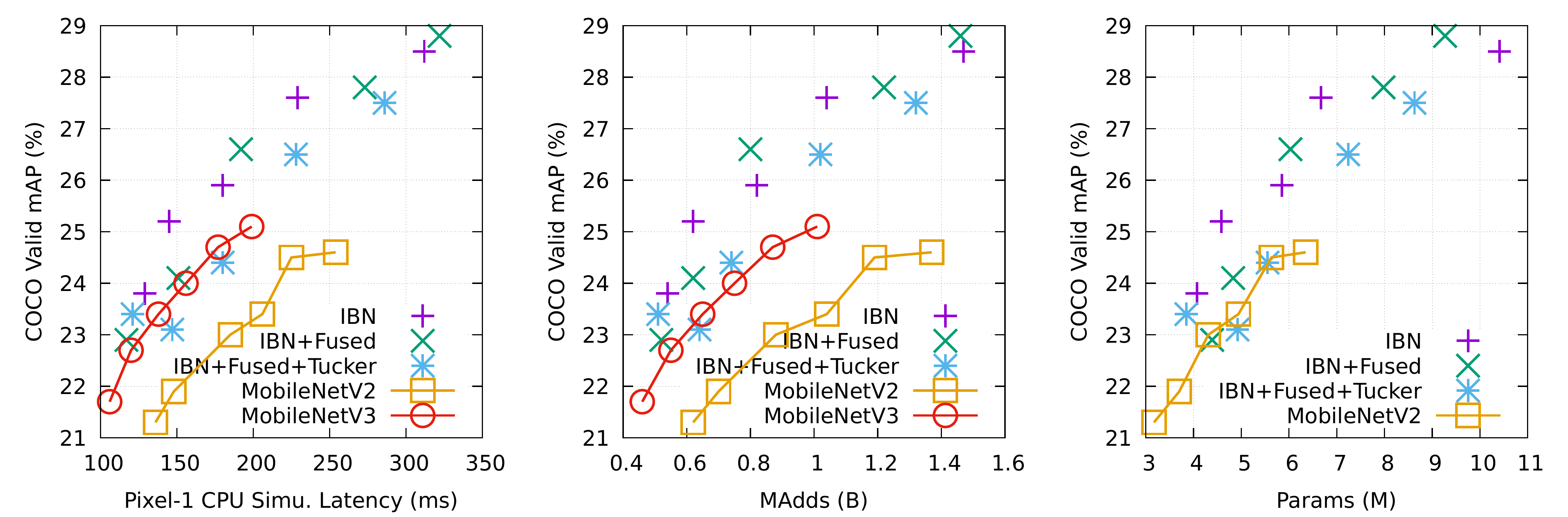}
    \caption{\footnotesize NAS results on Pixel-1 CPU using different search space variants.}
    \label{fig:search-space-ablation-cpu}
\end{figure*}

\begin{figure*}[!t]
    \centering
    \includegraphics[width=0.75\linewidth]{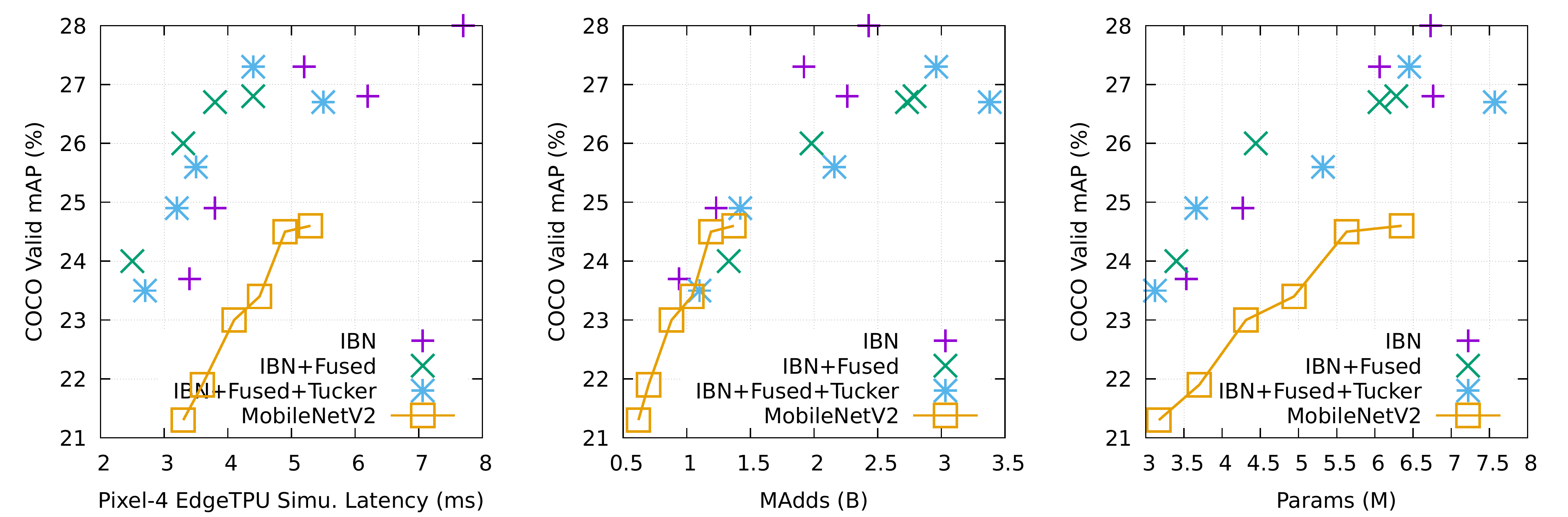}
    \caption{\footnotesize NAS results on Pixel-4 EdgeTPU using different search space variants.}
    \label{fig:search-space-ablation-edge-tpu}
\end{figure*}

\begin{figure*}[!t]
    \centering
    \includegraphics[width=0.75\linewidth]{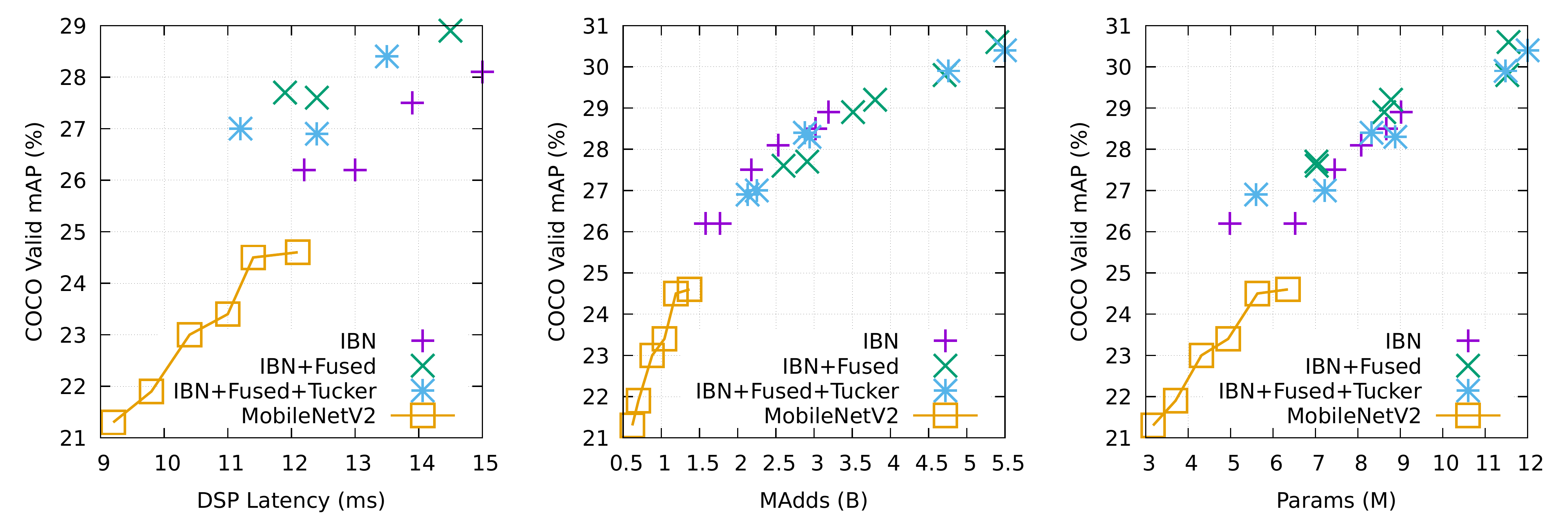}
    \caption{\footnotesize NAS results on Pixel-4 DSP using different search space variants.}
    \label{fig:search-space-ablation-dsp}
\end{figure*}

\smallskip
\noindent\textbf{CPU}. Figure~\ref{fig:search-space-ablation-cpu} shows the NAS results for Pixel-1 CPUs.
As expected, MobileNetV3+SSDLite is a strong baseline as the efficiency of its backbone has been heavily optimized for the same hardware platform over the classification task on ImageNet.
We also note that the presence of regular convolutions does not offer clear advantages in this particular case,
as IBN-only is already strong under FLOPS/CPU latency.
Nevertheless,
conducting domain-specific architecture search w.r.t. the object detection task offers non trivial gains on COCO (+1mAP in the range of 150-200ms).

\smallskip
\noindent\textbf{EdgeTPU}.
Figure~\ref{fig:search-space-ablation-edge-tpu} shows the NAS results when targeting Pixel-4 EdgeTPUs.
Conducting hardware-aware architecture search with any of the three search spaces significantly improves the overall quality. This is largely due to the fact that the baseline architecture (MobileNetV2)\footnote{MobileNetV3 is not well supported on EdgeTPUs due to h-swish and squeeze-and-excite blocks.} is heavily optimized towards CPU latency,
which is strongly correlated with FLOPS/MAdds but not well calibrated with the EdgeTPU latency.
Notably,
while IBN-only still offers the best accuracy-MAdds trade-off (middle plot),
having regular convolutions in the search space (either IBN+Fused or IBN+Fused+Tucker) offers clear further advantages in terms of accuracy-latency trade-off.
The results demonstrate the usefulness of full convolutions on EdgeTPUs.

\smallskip
\noindent\textbf{DSP}. 
Figure~\ref{fig:search-space-ablation-dsp} shows the search results for Pixel-4 DSPs. Similar to EdgeTPUs,
it is evident that the inclusion of regular convolutions in the search space leads to substantial mAP improvement under comparable inference latency.

\begin{table*}[hbt!]
\centering

\begin{threeparttable}
\resizebox{0.9\textwidth}{!}{
\begin{tabular}{@{}l|c|cc|ccc|cc@{}}
\toprule
\multirow{2}{*}{Model/Search Space} & Target & \multicolumn{2}{c}{mAP (\%)} & \multicolumn{3}{|c|}{Latency (ms)} & MAdds & Params \\
 & hardware & \hspace{0.2em}Valid & Test\hspace{0.2em} & \hspace{0.3em}CPU & \hspace{0.3em}EdgeTPU & \hspace{0.3em}DSP & (B) & (M) \\ \midrule
MobileNetV2$^{\Diamond\ddagger}$ &  &  -- & 22.1 & 162 & 8.4 & 11.3 & 0.80 & 4.3 \\
MobileNetV2 (ours)$^\Diamond$ &  & 22.2 & 21.8 & 129 & 6.5 & 9.2 & 0.62 & 3.17 \\
MobileNetV2 $\times$ 1.5 (ours)$^\Diamond$ & & 25.7 & 25.3 & 225 & 9.0 & 12.1 & 1.37 & 6.35 \\
MobileNetV3$^{\dagger\ddagger}$ &  & -- & 22.0 & 119 & $\ast$ & $\ast$ & 0.51 & 3.22 \\
MobileNetV3 (ours)$^\dagger$ &  & 22.2 & 21.8 & 108 & $\ast$ & $\ast$ & 0.46 & 4.03 \\
MobileNetV3 $\times$ 1.2 (ours)$^\dagger$ &  & 23.6 & 23.1 & 138 & $\ast$ & $\ast$ & 0.65 & 5.58 \\
MnasFPN (ours)$^\lhd$ & & 25.6 & 26.1 & 185 & 18.5 & 25.1 & 0.92 & 2.50 \\
MnasFPN (ours)$\times$ 0.7$^\lhd$ & & 24.3 & 23.8 & 120 & 16.4 & 23.4 & 0.53 & 1.29 \\
\midrule
IBN+Fused+Tucker$^\dagger$ & \multirow{3}{*}{CPU} & 24.2 & 23.7 & 122 & $\ast$ & $\ast$ & 0.51 & 3.85 \\
IBN+Fused$^\dagger$ &  & 23.0 & 22.7 & 107 & $\ast$ & $\ast$ & 0.39 & 3.57 \\
IBN$^\dagger$ &  & 23.9 & 23.4 & 113 & $\ast$ & $\ast$ & 0.45 & 4.21 \\ \midrule
IBN+Fused+Tucker & \multirow{3}{*}{\hspace{0.2em}EdgeTPU\hspace{0.2em}} & 25.7 & 25.5 & 248 &  6.9 & 10.8 & 1.53 & 4.20 \\
IBN+Fused &  & 26.0 & 25.4 & 272 & 6.8 & 9.9 & 1.76 & 4.79 \\
IBN &  & 25.1 & 24.7 & 185 & 7.4 & 10.4 & 0.97 & 4.17 \\ \midrule
IBN+Fused+Tucker$^\Diamond$ & \multirow{3}{*}{DSP} &  28.9 & 28.5 & 420 & 8.6 & 12.3 & 2.82 & 7.16 \\
IBN+Fused$^\Diamond$ &  & 29.1 & 28.5 & 469 & 8.6 & 11.9 & 3.22 & 9.15  \\
IBN$^\Diamond$ &  & 27.3 & 26.9 & 259 & 8.7 & 12.2 & 1.43 & 4.85 \\ \bottomrule
\end{tabular}
}
\end{threeparttable}
\vspace{-5pt}
\caption{\footnotesize Test AP scores are based on COCO test-dev. $\dagger$: Augmented with Squeeze-Excite and h-Swish (CPU-only); $\ast$: Not well supported by the hardware platform; $\Diamond$: 3$\times$3 kernel size only (DSP-friendly); $^\lhd$: Augmented with NAS-FPN head; $^\ddagger$: Endpoint C4 located after  the 1$\times$1 expansion in IBN.}
\label{tab:main-results}
\end{table*}

\subsection{Main Results}
We compare our architectures obtained via latency-aware NAS against state-of-the-art mobile detection models on COCO \cite{lin2014microsoft}.
For each target hardware platform,
we report results obtained by searching over each of the three search space variants. Results are presented in Table~\ref{tab:main-results}. On CPUs, we find that searching directly on a detection task allows us to improve mAP by 1.7mAP over MobileNetV3, but do not find evidence that our proposed fused IBNs and Tucker bottlenecks are required to obtain high quality models. On DSPs and EdgeTPUs, however, we find that our new primitives allow us to significantly improve the tradeoff between model speed and accuracy.

On mobile CPUs,
MobileDet outperforms MobileNetV3+SSDLite~\cite{howard2019searching},
a strong baseline based on the state-of-the-art image classification backbone,
by 1.7 mAP at comparable latency.
The result confirms the effectiveness of detection-specific NAS.
The models also achieved competitive results with MnasFPN, the state-of-the-art detector for mobile CPUs, without leveraging the NAS-FPN head which may complicate the deployment process.
It is also interesting to note that the incorporation of full convolutions is quality-neutral over mobile CPUs,
indicating that IBNs are indeed promising building blocks for this particular hardware platform. 

On EdgeTPUs, MobileDet outperforms MobileNetV2+SSDLite by 3.7 mAP on COCO test-dev at comparable latency.
We attribute the gains to both task-specific search (w.r.t. COCO and EdgeTPU) and the presence of full convolutions.
Specifically,
IBN+Fused+Tucker leads to 0.8 mAP improvement together with $7\%$ latency reduction as compared to the IBN-only search space.

On DSPs, MobileDet achieves $28.5$ mAP on COCO with $12.3$ ms latency, outperforming MobileNetV2+SSDLite ($\times 1.5$) by $3.2$ mAP at comparable latencies.
The same model also outperforms MnasFPN by $2.4$ mAP with more than $2 \times $ speed-up.
Again it is worth noticing that including full convolutions in the search space clearly improved the architecture search results from 26.9 mAP @ 12.2 ms to 28.5 mAP @ 11.9ms.

These results support the fact that mobile search spaces have traditionally targeted CPU, which heavily favors separable convolutions, and that we need to rethink this decision for networks targeting other hardware accelerators. 

\begin{figure}[!hbtp]
    \centering
    \includegraphics[width=0.95\linewidth]{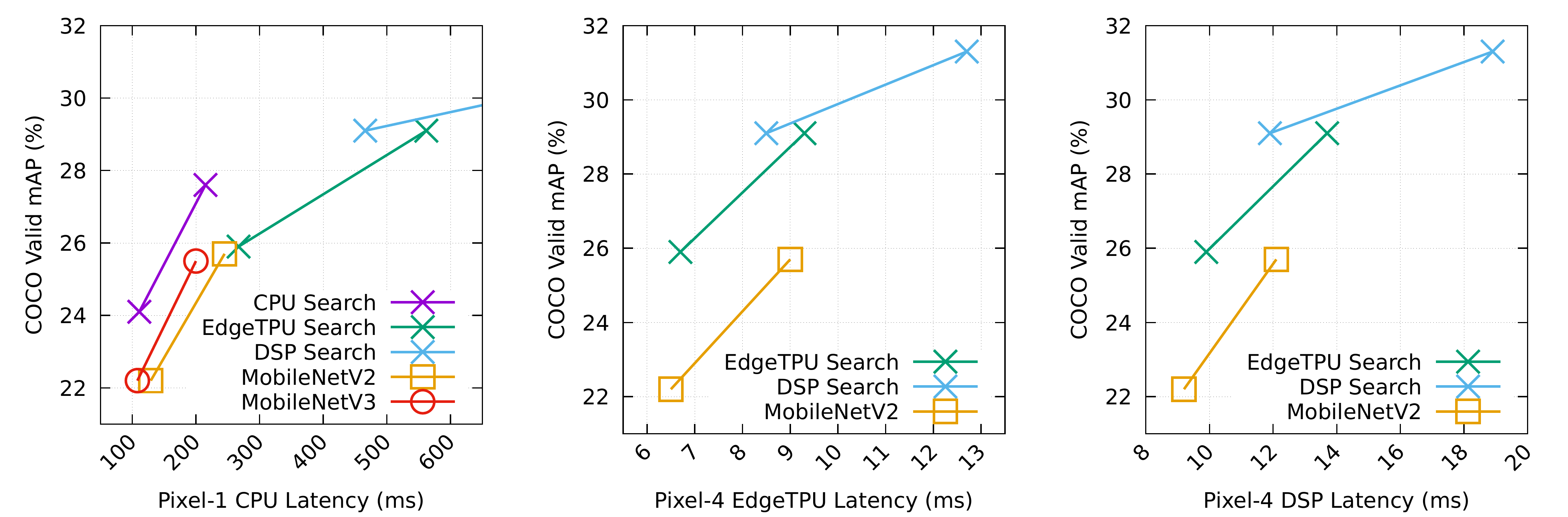}
    \caption{\footnotesize Transferability of architectures (searched for   different target platforms) across hardware platforms. For each given architecture,
    we report the original model and its scaled version with channel multiplier 1.5$\times$.}
    \label{fig:transferability}
    \vspace{-5pt}
\end{figure}

\label{sec:model-visualization}
\begin{figure*}[h!]
\centering
\begin{minipage}[t]{0.24\textwidth}
\centering Target: Pixel-1 CPU \\
(23.7 mAP @ 122 ms)
  \includegraphics[width=0.75\linewidth]{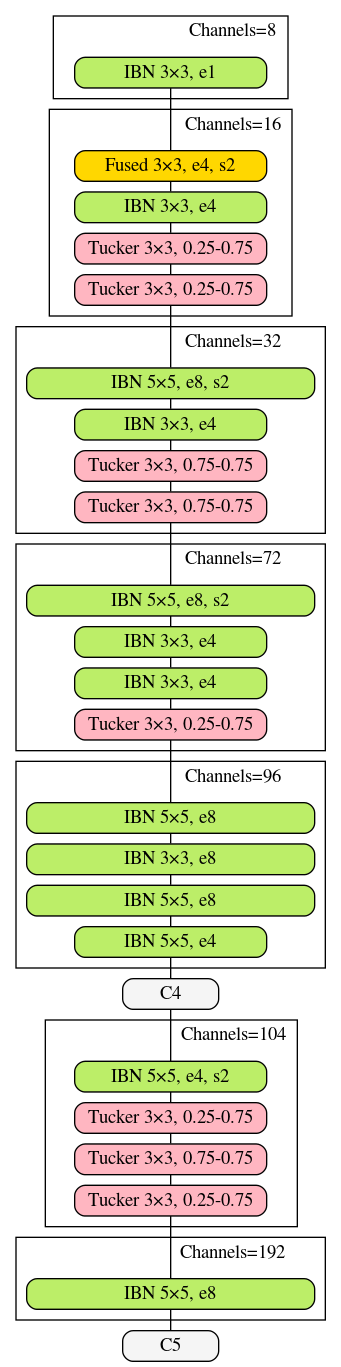}
\end{minipage}
\begin{minipage}[t]{0.24\textwidth}
\centering Target: Pixel-4 EdgeTPU \\
(25.5 mAP @ 6.8 ms)
  \includegraphics[width=0.75\linewidth]{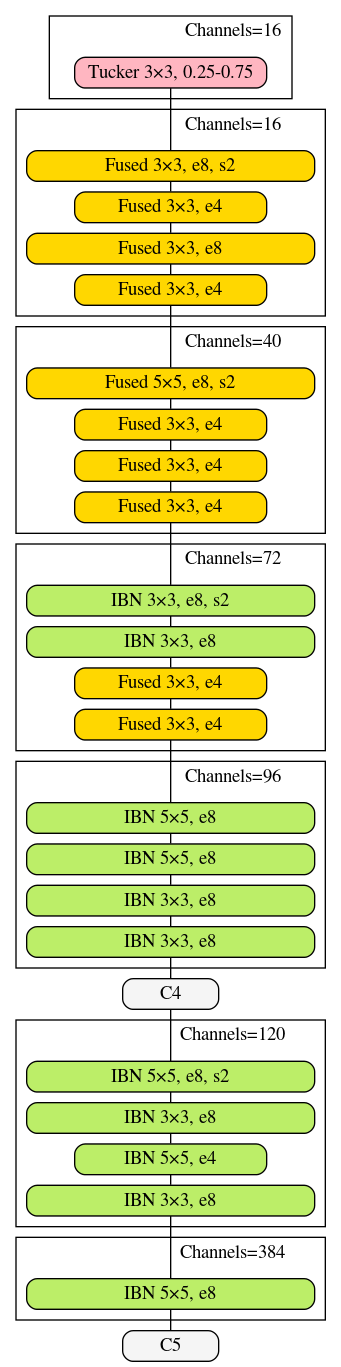}
\end{minipage}
\begin{minipage}[t]{0.24\textwidth}
\centering Target: Pixel-4 DSP \\
(28.5 mAP @ 12.3 ms)
  \includegraphics[width=0.75\linewidth]{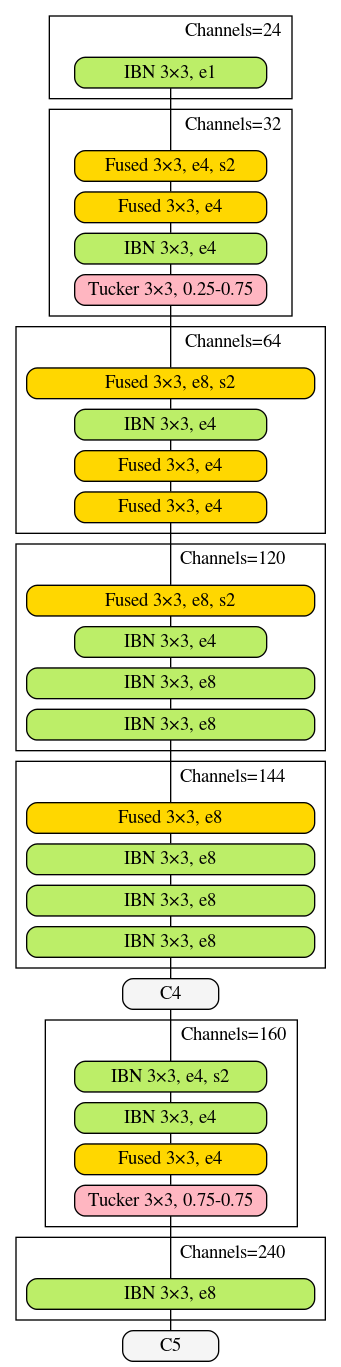}
\end{minipage}
\begin{minipage}[t]{0.24\textwidth}
\centering Target: Jetson Xavier GPU \\
(28.0 mAP @ 3.2 ms)
  \includegraphics[width=0.75\linewidth]{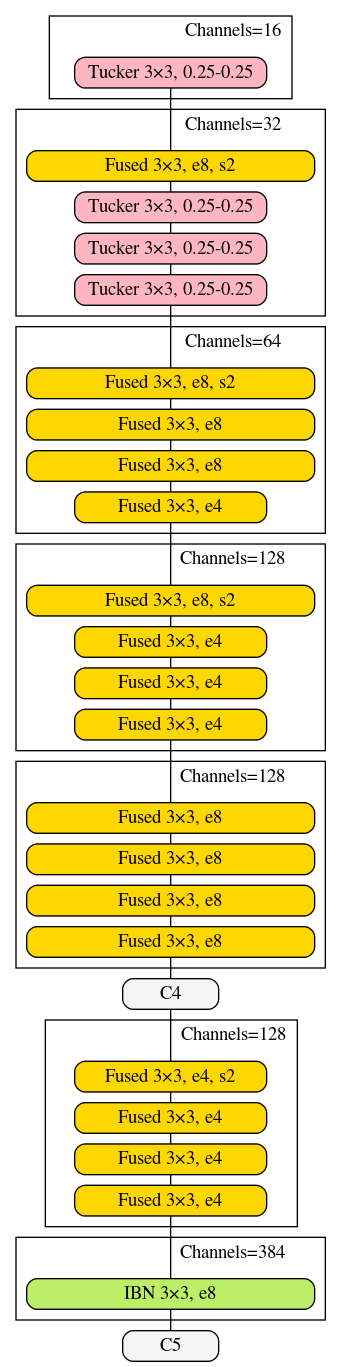}
\end{minipage}
\vspace{-10pt}
\caption{\footnotesize Best architectures searched in the IBN+Fused+Tucker space w.r.t. different mobile accelerators.
Endpoints C4 and C5 are used by the SSDLite head. In the figures above, $e$ refers to the expansion factor, $s$ refers to the stride, and ``Tucker 3$\times$3, 0.25-0.75'' refers to a Tucker layer with kernel size $3\times 3$, input compression ratio 0.25 and output compression ratio 0.75.}
\label{fig:mobiledets}
\end{figure*}

\smallskip
\noindent\textbf{Cross-Hardware Transferability of Searched Architectures}. Fig.~\ref{fig:transferability} compares MobileDets (obtained by targeting at different accelerators) w.r.t. different hardware platforms. Our results indicate that architectures searched on EdgeTPUs and DSPs are mutually transferable. In fact, both searched architectures extensively leveraged regular convolutions. On the other hand, architectures specialized w.r.t.  EdgeTPUs or DSPs (which tend to be FLOPS-intensive) do not transfer well to mobile CPUs.

\smallskip
\noindent\textbf{Generalization to New Hardware}. 
The proposed search space family was specifically developed for CPUs, EdgeTPUs and DSPs.
It remains an interesting question whether the search space can extrapolate to new hardware. To answer this, we conduct architecture search for NVIDIA Jetson GPUs as a ``holdout'' device.
We follow the DSP setup in Table~\ref{tab:main-results}, except that the number of filters are rounded to the multiples of 16.
Results are reported in Table~\ref{tab:gpu-results}.
The searched model within our expanded search space achieves +3.7mAP boost over MobileNetV2 while being faster.
The results confirm that the proposed search space family is generic enough to handle a new hardware platform.

\begin{table}
    \centering
    \small
    \resizebox{0.48\textwidth}{!}{
    \begin{tabular}{@{}c|cc|c@{}}
    \toprule
        \multirow{2}{*}{Model} & \multicolumn{2}{c|}{mAP (\%)} & Jetson Xavier \\
        & Valid & Test & FP16 Latency (ms) \\
        \midrule
        MobileNetV2 (ours) & 22.2 & 21.8 & 2.6 \\
        MobileNetV2 $\times$ 1.5 (ours) & 25.7 & 25.3 & 3.4 \\
        Searched (IBN+Fused+Tucker) & 27.6 & 28.0 & 3.2 \\
        \bottomrule
    \end{tabular}}
    \vspace{-0.7em}
    \caption{\footnotesize Generalization of the MobileDet search space family to unseen hardware (edge GPU).}
    \label{tab:gpu-results}
    \vspace{-10pt}
\end{table}

\smallskip
\noindent\textbf{Architecture visualizations}.  Fig.~\ref{fig:mobiledets} illustrates our searched object detection architectures, MobileDets, by targeting at different mobile hardware platforms using our largest search space. One interesting observation is that MobileDets use regular convolutions extensively on EdgeTPU and DSP, especially in the early stage of the network where depthwise convolutions tend to be less efficient. The results confirm that IBN-only search space is not optimal for these accelerators. 

\section{Conclusion}
In this work, we question the predominant design pattern of using depthwise inverted bottlenecks as the only building block for vision models on the edge. Using the object detection task as a case study,
we revisit the usefulness of regular convolutions over a variety of mobile accelerators, including mobile CPUs, EdgeTPUs, DSPs and edge GPUs. Our results reveal that full convolutions can substantially improve the accuracy-latency trade-off on several accelerators when placed at the right positions in the network, which can be efficiently identified via neural architecture search. The resulting architectures, MobileDets, achieve superior detection results over a variety of hardware platforms, significantly outperforming the prior art by a large margin.

\medskip

{\small
\bibliographystyle{ieee_fullname}
\bibliography{main}
}

\newpage
\appendix
\section{Appendix}
In this appendix, we  describe further details of the relationship between the building blocks of MobileDet search space and the linear structure of Tucker/CP decomposition.
\subsection{Connections with Tucker/CP decomposition}
The proposed layer variants can be linked to Tucker/CP decomposition. Fig. ~\ref{fig:cp_layers} shows the graphical structure of an inverted bottleneck with input expansion ratio $s$, modulo nonlinearities. This structure is equivalent to the sequential structure of approximate evaluation of a regular convolution by using CP decomposition \cite{lebedev2014speeding}. The Tucker convolution layer with input and output compression ratios $s$ and $e$, denoted as Tucker layer shown in Fig. ~\ref{fig:tucker_layers}, has the same structure (modulo nonlinearities) as the Tucker decomposition approximation of a regular convolution~\cite{kim2015compression}. Fused inverted bottleneck layer with an input expansion ratio $s$, shown in Fig. ~\ref{fig:fused_layers}, can also be considered as a variant of the Tucker decomposition approximation.

Details of the approximation is as follows. CP-decomposition approximates convolution using a set of sequential linear mappings: a $1\times 1$ pointwise convolution, two depthwise convolutions along the spatial dimensions, and finally another pointwise convolution. Since the kernel size of convolution is quite small, e.g. 3$\times$3, or 5$\times 5$, the decomposition along the spatial dimensions does not save much computation. Without performing the decomposition along the spatial dimensions, the sequential graphical structure is equivalent to inverted bottleneck in Fig. \ref{fig:cp_layers}. Similarly, a mode-2 Tucker decomposition approximation of a convolution along input and output channels involves a sequence of three operations, a $1 \times 1$ convolution, then a $K\times K$ regular convolution, then another $1\times 1$ pointwise convolution. This sequential structure of Tucker layer is shown in Fig.~\ref{fig:tucker_layers}. Combining the first $1\times 1$ pointwise convolution and the second $K \times K$ regular convolution as one $K \times K$ regular convolution gives the fused inverted bottleneck layer in Fig.~\ref{fig:fused_layers}.

We therefore refer to the expansion operation as fused convolution layer, the compression operation as Tucker layer, and our proposed search space with a mix of both layers and IBNs as the MobileDets search space. 

\end{document}